\pdfoutput=1

\documentclass[11pt]{article}

\PassOptionsToPackage{dvipsnames}{xcolor}

\usepackage{acl}

\usepackage{times}
\usepackage{latexsym}

\usepackage[T1]{fontenc}

\usepackage[utf8]{inputenc}

\usepackage{microtype}

\usepackage{inconsolata}

\usepackage{graphicx}

\usepackage{natbib}
\usepackage{bbm}
\usepackage{amsmath,amssymb,amsfonts}
\usepackage{subfigure}
\usepackage{algorithmic}
\usepackage{graphicx}
\usepackage{textcomp}
\usepackage{tabularx}
\usepackage{hyperref}
\usepackage{multirow}
\usepackage{caption}
\usepackage{booktabs} 
\usepackage{colortbl}
\usepackage{subcaption}
\usepackage{rotating}
\usepackage{wrapfig}
\usepackage{tikz}
\usepackage{cleveref}
\usepackage{amsthm} 
\newtheorem{lemma}{Lemma}

\title{FocalPO: Enhancing Preference Optimizing by Focusing on Correct Preference Rankings}
\author{Tong Liu$^{1,5}$, \ Xiao Yu$^{2 \dag}$, \ Wenxuan Zhou$^{3}$, \ Jindong Gu$^{4}$, \ Volker Tresp$^{1,5}$ \\
$^1$LMU Munich \quad $^2$Columbia University \quad $^3$University of Southern California 
\\ 
$^4$University of Oxford \quad $^5$Munich Center for Machine Learning 
\\
\texttt{tongliu.physics@gmail.com} \quad 
\texttt{xy2437@columbia.edu} }

\begin{document}
\maketitle

\begin{abstract}
Efficient preference optimization algorithms such as Direct Preference Optimization (DPO) have become a popular approach in aligning large language models (LLMs) with human preferences.
These algorithms implicitly treat the LLM as a reward model, and focus on training it to correct misranked preference pairs.
However, recent work~\citep{chen2024preference} empirically finds that DPO training \textit{rarely improves these misranked preference pairs}, despite its gradient emphasizing on these cases. 
We introduce FocalPO, a DPO variant that instead \textit{down-weighs} misranked preference pairs and prioritizes enhancing the model's understanding of pairs that it can already rank correctly.
Inspired by Focal Loss used in vision tasks, FocalPO achieves this by adding a modulating factor to dynamically scale DPO loss. 
Our experiment demonstrates that FocalPO surpasses DPO and its variants on popular benchmarks like Alpaca Eval 2.0 and Arena-Hard using Mistral-Base-7B and Llama-3-Instruct-8B, with the introduced hyperparameter fixed. 
Additionally, we empirically reveals how FocalPO affects training on correct and incorrect sample groups, further underscoring its effectiveness~\footnote{Code available at:~\url{https://github.com/TongLiu-github/focalpo}}.


\end{abstract}

\section{Introduction}

Reinforcement learning from human feedback (RLHF) played a crucial role in aligning large language models (LLMs) with human preferences~\citep{Christiano_RLHF, Ouyang_RLHF, Stiennon_RLHF}. 
However, conducting RLHF with Proximal Policy Optimization (PPO, \citealt{schulman2017proximal}) is computationally expensive.
Therefore, recent works have studied more efficient approaches, such as Direct Preference Optimization (DPO, \citealt{Rafailov_DPO}) and its variants \citep{AminiVC24ODPO,simpo,KTO,zhou-etal-2024-wpo}, which implicitly treat the language model itself as a reward model and directly optimizes it using preference datasets.

Despite its popularity, recent work~\cite{chen2024preference} shows that DPO often \emph{fails to correct incorrect preference rankings made by the implicit reward model prior to training}, despite its gradient emphasizing these cases.
While one recent competitive DPO variant SimPO~\citep{simpo} introduces a fixed margin term in the preference optimization loss, we find it \emph{inherently reduces} the disproportionately high weights assigned to learning incorrect response pairs. 
ODPO~\cite{AminiVC24ODPO} further incorporates additional labeled reward values and employs an instance-dependent margin term. 
However, while these approaches implicitly re-weigh instances through the use of margin terms, they lack explicit and dynamic mechanisms for such re-weighing.

\begin{figure}[t]
\centering 
\hspace{-5mm}
\hspace{-5mm}
\subfigure{
\includegraphics[width=0.4\textwidth]{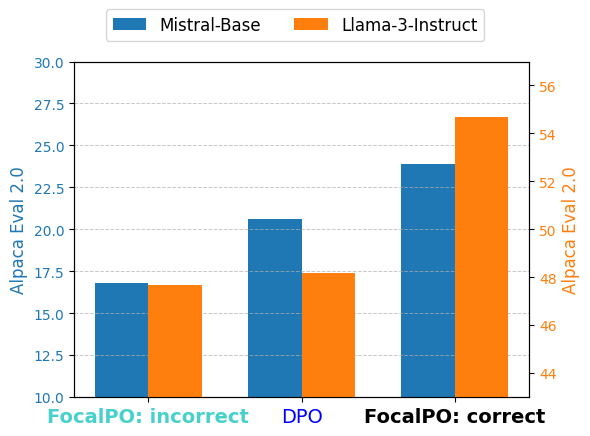}}
\hspace{-5mm}
\hspace{-5mm}
\vspace{-2mm}\\
\caption{
Comparing the effectiveness of up-weighing \textbf{\textcolor{BlueGreen}{incorrect}} preference pairs against up-weighing \textbf{correct} preference pairs with FocalPO during training.
We find that performance further improves by prioritizing learning from pairs that can already be ranked correctly.
}
\label{fig:focus_correct_incorrect}
\vspace{-2mm}
\end{figure}

To this end, we propose FocalPO, a loss function that \emph{dynamically and explicitly reduces} DPO's emphasis on learning response pairs that the model struggles to rank while prioritizing preference pairs with accurate implicit reward estimates.
Inspired by \textit{Focal Loss}~\citep{LinGGHD20} in vision tasks, we achieve this by adding a modulating factor of $p^{\gamma}$ to dynamically scale the DPO loss, $-\log p$. 
Here, $p$ represents the probability that the preferred response over the dispreferred one.
Intuitively, such a factor automatically up-weighs the contribution of correctly ranked samples during training, with the scaling factor decays to zero as preference probability increases. 
Gradient analysis shows that FocalPO equivalently assigns higher weights to pairs whose implicit reward estimate order is accurate and lower weights to incorrect pairs.
This encourages models to learn from preference pairs with correct reward estimates while avoiding excessive influence from challenging ones.
Experiments show that FocalPO significantly outperforms the original DPO.
Additionally, we provide case study showing how FocalPO affects training on correct and incorrect sample groups, further underscoring its effectiveness.

Our contributions are summarized as follows: 
    (1) We introduce FocalPO, a novel loss function based on DPO and Focal Loss, which emphasizes learning preference pairs with accurate implicit reward estimates.
    (2) We show that FocalPO surpasses DPO as well as many DPO variants on popular chat benchmarks like Alpaca-Eval 2.0 and Arena-Hard, using both Mistral-Base-7B and Llama-3-Instruct-8B models.
    (3) We empirically demonstrate the impact of prioritizing learning correctly vs. incorrect ranked preference pairs.

\section{Background} 
Given an instruction-tuned language model $\pi_{\theta}$, Direct Preference Optimization (DPO, \citet{Rafailov_DPO}) further optimizes it as an implicit reward model using a preference dataset.
Let $\mathcal{D} = \{(x^{(i)}, y^{(i)}_{w}, y^{(i)}_{l})\}_{i}^{N}$ denote a preference dataset where $x^{(i)} \in \mathcal{X}$ is a prompt and $(y^{(i)}_{w}, y^{(i)}_{l})$ is a pair of answers, with the preference of $y^{(i)}_{w} \succ y^{(i)}_{l}$ expressed by human or AI annotators.
Under the assumption of Bradley-Terry model~\citep{bradley1952rank}, the human preference distribution $p^{*}$ can be approximated as: 
\begin{equation}
\scalebox{0.88}{$
\begin{aligned}
p^*(y^{(i)}_{w} \succ y^{(i)}_{l} \mid x^{(i)}) = \sigma(r^{*}(x^{(i)}, y^{(i)}_{w}) - r^{*}(x^{(i)}, y^{(i)}_{l})),
\end{aligned}$}
\end{equation} 
where $\sigma$ and $r^{*}(x, y)$ are the sigmoid function and the latent reward model used to generate the ground-truth preference, respectively.
To model this human preference, DPO uses a reparametrization trick to express it in terms of the optimal policy $\pi^{*}$: $r^{*}(x, y)=\beta \log \frac{\pi^{*}(y \mid x)}{\pi_{\text {ref }}(y \mid x)}+\beta \log Z(x)$, where $Z(x)$ is the partition function only based on $x$.
Applying the above equations, the maximum likelihood estimation objective of DPO is: 
\begin{equation}
\scalebox{0.9}{$
\begin{aligned}
\mathcal{L}&_{\mathrm{DPO}}\left(\pi_\theta ; \pi_{\mathrm{ref}}\right) 
=-\mathbb{E}_{\left(x, y_w, y_l\right) \sim \mathcal{D}}
\left[ \log p\left(y_{w} \succ y_{l} \mid x \right) \right] \\
&=-\mathbb{E}_{\left(x, y_w, y_l\right) \sim \mathcal{D}}
\left[ \log \sigma \left( \beta \log \frac{\pi_\theta\left(y_w \mid x\right)}{\pi_{\mathrm{ref}}\left(y_w \mid x\right)}\right.\right. \\
&\quad\left.\left.- \beta \log \frac{\pi_\theta\left(y_l \mid x\right)}{\pi_{\mathrm{ref}}\left(y_l \mid x\right)} \right)\right], 
\label{eq_dpoloss}
\end{aligned}$}
\end{equation}
which \emph{emphasizes} on learning responses pairs that the model struggle to rank (i.e., wrong reward estimates).
However, recent work such as \citet{chen2024preference} shows that DPO cannot learn to correct these response pairs, and a few other competitive variants such as SimPO~\citep{simpo} and ODPO~\citep{AminiVC24ODPO} \emph{implicitly introduce a margin term} that reduces such emphasis.

\section{FocalPO: Focal loss inspired Preference Optimization}

\subsection{FocalPO loss} 
\begin{figure}[t]
\centering 
\hspace{-10mm}
\hspace{-5mm}
\subfigure{
\includegraphics[width=0.48\textwidth]{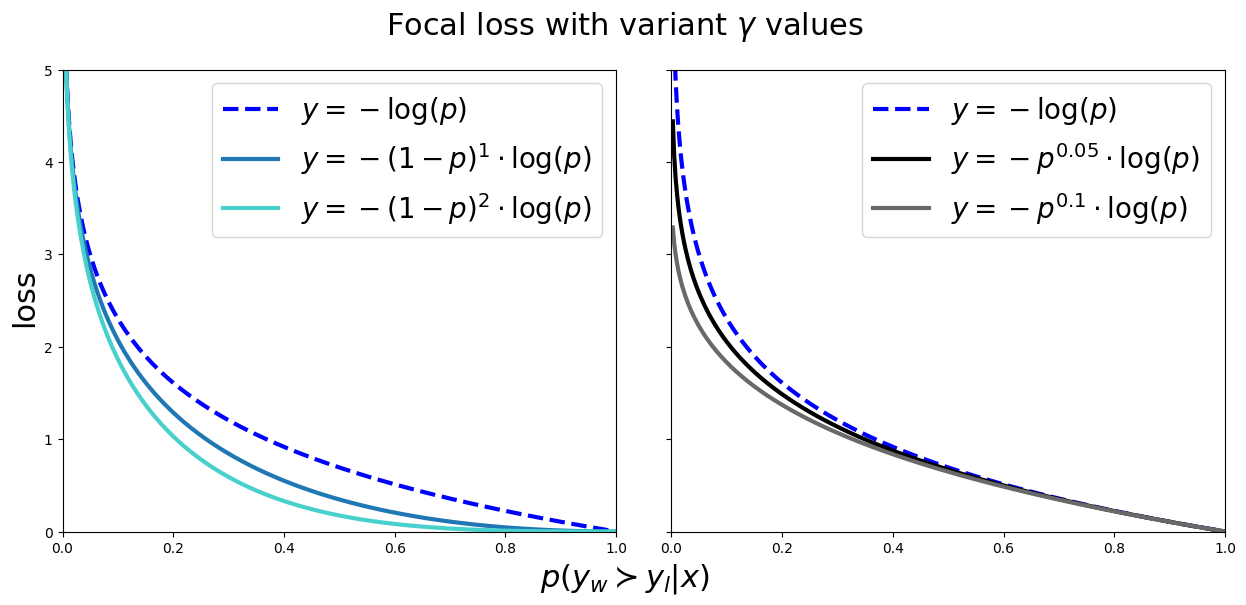}}
\hspace{-5mm}
\hspace{-10mm}
\vspace{-2mm}\\
\caption{We scale DPO loss using \textit{Focal Loss}. Unlike the original Focal Loss (left), we add a scaling factor of $p^{\gamma}$ (right) to the dynamically adjust the cross entropy loss. This modification enables a sample-wise adjustment of weights, reducing the influence of high weights when the implicit reward model order is incorrect.} 
\label{fig:focal_demo}
\end{figure}

We propose FocalPO loss, which directly scales DPO's loss to reduce emphasis on learning response pairs the model struggles to rank and focuses on preference pairs with accurate implicit reward estimates.
Inspired by Focal Loss \citep{LinGGHD20} in vision tasks (i.e., adding a modulating factor $(1-p)^{\gamma}$ to the cross entropy loss), we replace the original cross-entropy loss in Eq.~\ref{eq_dpoloss} with: 
\begin{equation}
\scalebox{0.88}{$
\begin{aligned}
&\mathcal{L}_{\mathrm{FocalPO}}\left(\pi_\theta ; \pi_{\mathrm{ref}}\right) 
= -\mathbb{E}_{\left(x, y_w, y_l\right) \sim \mathcal{D}}\\ 
&\left[\big( 1 - p\left(y_{w} \succ y_{l} \mid x \right) \big)^{-\gamma}\cdot \log p\left(y_{w} \succ y_{l} \mid x \right)\right]
\end{aligned}$}
\label{eq_focalpoloss1}
\end{equation}
\begin{equation}
\scalebox{0.88}{$
\begin{aligned}
&\approx -\mathbb{E}_{\left(x, y_w, y_l\right) \sim \mathcal{D}}
\left[p\left(y_{w} \succ y_{l} \mid x \right)^{\gamma} \log p\left(y_{w} \succ y_{l} \mid x \right)\right] \\
&\quad + \text{constant},
\end{aligned}$}
\label{eq_focalpoloss2}
\end{equation}
where $\gamma \geq 0$ is a tunable \textit{focusing} hyperparameter. 
Here we take negative index $-\gamma$ to replace the original positive term $\gamma$ in Focal Loss, in order to focus on the easier or correctly ranked preference pairs. 
To simplify the optimization process, we further approximate $(1-p)^{-\gamma} \log p$~\footnote{In experiments, we also observe that training with the power of $-\gamma$ often causes the gradient to become NaN. We reuse the symbol $\gamma$ for notational simplicity, although the exponent arises from an approximation. } with the term $p^{\gamma} \log p$. 
Unless otherwise specified, we use Eq.~\ref{eq_focalpoloss2} as FocalPO for the rest of the paper. Figure~\ref{fig:focal_demo} illustrates the focal loss with modulating factors $(1-p)^{\gamma}$ and $p^{\gamma}$.
Converse to $(1-p)^{\gamma}$, a higher $\gamma$ in $p^{\gamma}$ leads to a less emphasis placed on hard negative samples.
When the implicit reward model "correctly classifies" a preference pair (i.e., $p(y_{w} \succ y_{l}|x)$ is large), the modulating factor approaches 1, leaving the loss \textit{almost unchanged}.
Conversely, as $p(y_{w} \succ y_{l}|x) \rightarrow 0$, the modulating factor approaches 0, effectively \textit{down-weighing} the preference pair in the DPO loss.

\subsection{Gradient analysis} 
To better understand the impact of modulating factor on the DPO loss, we perform a gradient analysis.
The gradient of DPO with respect to $\theta$ is:  
\begin{equation}
\scalebox{0.9}{$
\begin{aligned}
\nabla_{\theta} \mathcal{L}_{\mathrm{DPO}} 
= -\beta \mathbb{E}_{\left(x, y_w, y_l\right) \sim \mathcal{D}} 
\bigg[ & \textcolor{blue}{\sigma\big(\hat{r}_{\theta}(y_l) - \hat{r}_{\theta}(y_w)\big)} \\
& \cdot \nabla_{\theta} \frac{\log \pi(y_{w}|x)}{\log \pi(y_{l}|x)} \bigg],
\end{aligned}$}
\label{eq_dpogradient}
\end{equation}
where $\hat{r}_{\theta}(y)=\beta \log \frac{\pi_{\theta}(y|x)}{\pi_{ref}(y|x)}$ is the reward implicitly defined by the language model $\pi_{\theta}$ and reference model $\pi_{ref}$. 
The term $\sigma(\hat{r}_{\theta}(y_l)-\hat{r}_{\theta}(y_w))$ assigns higher weights when reward estimate is wrong. 

\begin{figure}[t]
\centering 
\hspace{-7mm}
\hspace{-7mm}
\subfigure{
\includegraphics[width=0.4\textwidth]{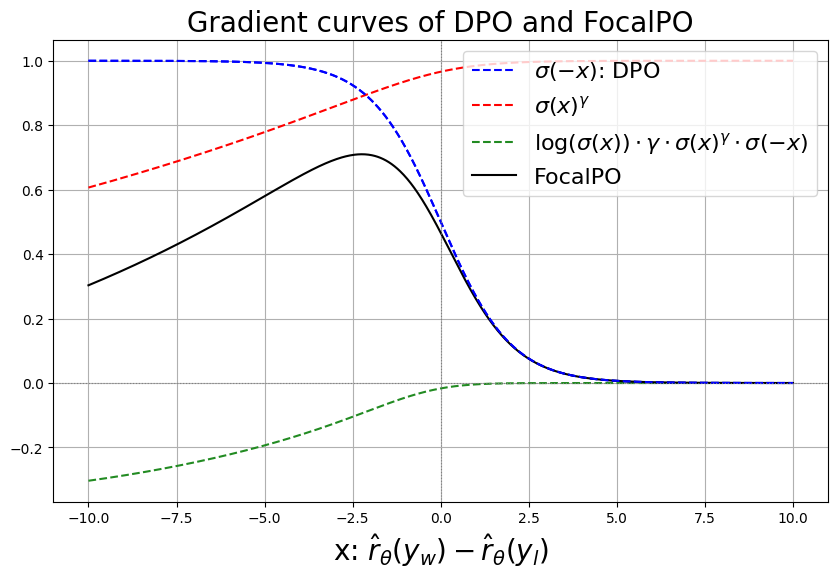}}
\hspace{-7mm}
\hspace{-7mm}
\vspace{-2mm}\\
\caption{Gradients of DPO and FocalPO. We used $\gamma=0.05$ for $p^{\gamma}$ in FocalPO in this example.}
\label{fig:gradient}
\end{figure}

In comparison, the gradient of FocalPO~\footnote{Refer to Appendix~\ref{appx:gradient} for detailed derivation and Appendix~\ref{appx:relation} for an interpretation of FocalPO.} is: 
\begin{equation}
\scalebox{0.9}{$
\begin{aligned}
\nabla_{\theta} \mathcal{L}&_{\mathrm{FocalPO}} 
= -\beta \mathbb{E}_{\left(x, y_w, y_l\right) \sim \mathcal{D}} \left[ \left[ \textcolor{blue}{\sigma(\hat{r}_{\theta}(y_l)-\hat{r}_{\theta}(y_w))} \right. \right. \\
& \cdot \textcolor{red}{\sigma^{\gamma}(\hat{r}_{\theta}(y_w)-\hat{r}_{\theta}(y_l))} + \textcolor{ForestGreen}{\log \sigma(\hat{r}_{\theta}(y_w)-\hat{r}_{\theta}(y_l))} \\ 
& \left. \cdot \textcolor{ForestGreen}{\gamma \sigma^{\gamma}(\hat{r}_{\theta}(y_w)-\hat{r}_{\theta}(y_l))} \cdot \textcolor{ForestGreen}{\sigma(\hat{r}_{\theta}(y_l)-\hat{r}_{\theta}(y_w))} \right] \\
& \left. \cdot \nabla_{\theta} \frac{\log \pi(y_{w}|x)}{\log \pi(y_{l}|x)} \right].
\end{aligned}$}
\label{eq_fpogradient}
\end{equation}
We illustrate the influence of each term in Fig.~\ref{fig:gradient}.
The blue curve represents the gradient term from DPO, while the red and green curves correspond to the additional gradient terms introduced by our modulating factor.
Compared to DPO, the new gradient terms from FocalPO assign lower weights to incorrect preference pairs.
When combined, as shown in the black curve in Fig.~\ref{fig:gradient}, FocalPO forms \textit{a bell-shaped curve}, assigning the highest weights to preference pairs with reward margins close to zero while assigning lower weights for both correctly classified pairs and significantly incorrect pairs.
FocalPO thus aims to mitigate the impact of preference pairs where reward estimates strongly disagree with the preference labels, and shifting focus toward learning preference pairs the model already ranks correctly.

\begin{table*}[h]
\centering
\begin{tabular}{lcccccc}
\hline
& \multicolumn{3}{c}{Llama-3-Instruct (8B)} & \multicolumn{3}{c}{Mistral-Base (7B)} 
\\ \cline{2-7} 
Models & \multicolumn{2}{l}{Alpaca Eval 2.0} & Arena-Hard & \multicolumn{2}{l}{Alpaca Eval 2.0} & Arena-Hard
\\ \cline{2-7} 
& WR & LCWR & WR & WR & LCWR & WR
\\ \hline
ORPO & 33.8 & 38.1 & 26.0 & 12.6 & 14.7 & 6.2 
\\
KTO & 31.8 & 33.1 & 11.7 & 12.3 & 14.9 & 8.8 
\\
DPO & 47.5 & 48.2 & 33.1 & 18.6 & 20.6 & 16.4 
\\
SimPO & 47.5 & 53.7 & 33.8 & \textbf{21.4} & 21.5 & 17.0 
\\
FocalPO & \textbf{49.8} & \textbf{54.7} & \textbf{34.6} & 20.1 & \textbf{22.5} & \textbf{17.1}
\\ \hline
\end{tabular}
\caption{Alpaca Eval 2.0 and Arena-Hard results of DPO, its variants and FocalPO for preference optimazation of Mistral-Base (7B) and Llama-3-Instruct (8B). }
\label{tab:main_result}
\end{table*}


\section{Experimental setup} 
\paragraph{Model and training datasets} 
We perform preference optimization on two representative models, Mistral-Base SFT (7B)~\footnote{\url{https://huggingface.co/HuggingFaceH4/mistral-7b-sft-beta}} and instruction-tuned Llama-3 (8B)~\footnote{\url{https://huggingface.co/meta-llama/Meta-Llama-3-8B-Instruct}}. We perform perference learning on the UltraFeedback dataset~\citep{UltraFeedback} \footnote{\url{https://huggingface.co/datasets/trl-lib/ultrafeedback_binarized}} for Mistral-Base, and on the Llama3-ultrafeedback-armorm dataset~\footnote{\url{https://huggingface.co/datasets/princeton-nlp/llama3-ultrafeedback-armorm}} for Llama-3-Instruct. 
The former dataset is sampled from multiple LLMs and judged by GPT-4, and the later contains preference pairs generated by Llama-3-Instruct and judged by ArmoRM~\citep{wang2024interpretable}. 

\paragraph{Hyperparameters}

For Mistral-Base (7B), we adopt the official hyperparameters from Zephyr~\citep{zephyr} and use $\beta=0.01$, epoch=1, batch size as 128 and learning rate as 5e-7.
For instruction-tuned Llama-3 (8B), we follow the setting of SimPO \citep{simpo}.
We use a batch size of 128, epoch=1, and for each algorithm, we report the best performance after a grid searching learning rates in [3e-7, 5e-7, 6e-7, 1e-6]. 
We also fix the hyperparameter of $\gamma$ in FocalPO as $0.05$ for both models to prevent possible overfitting of hyperparameters.

\paragraph{Baselines}
We compare FocalPO against DPO, as well as other popular DPO variants including KTO~\citep{KTO}, ORPO~\citep{hong-etal-2024-orpo}, and SimPO~\citep{simpo}.

\paragraph{Evaluation}
To compare the effectiveness of preference learning, we use Alpaca Eval 2.0~\citep{alpaca_eval,dubois2024length} and Arena-Hard~\citep{li2024crowdsourced}. 
Alpaca Eval 2.0 includes 805 representative instructions drawn from multiple datasets and compares model's responses with reference responses generated by \texttt{gpt-4-turbo}. 
We use default \texttt{gpt-4-turbo} as the judge model. 
We present metrics of the win rate (WR) and the length-controlled win rate (LCWR). 
Arena-Hard contains 500 well-defined challenging prompts and compares model's responses with reference responses generated by \texttt{gpt-4-0314}. 
We use default \texttt{gpt-4-1106-preview} as the judge model. 
We present the metric of WR. 
Both benchmarks are evaluated under the standard setting. 

\vspace{-1mm}
\section{Results} 
\vspace{-1mm}

\begin{figure}[t]
\centering 
\hspace{-5mm}
\hspace{-5mm}
\vspace{-4mm}
\subfigure{
\includegraphics[width=0.4\textwidth]{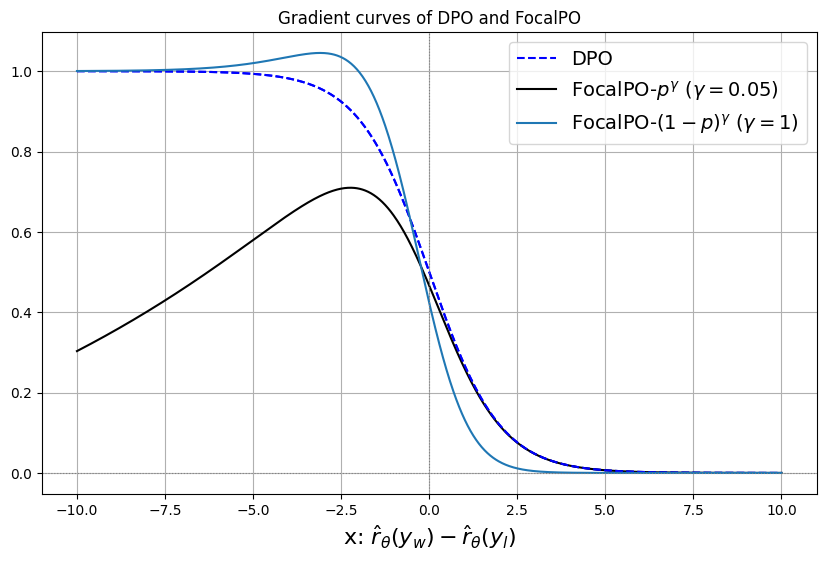}}
\hspace{-5mm}
\hspace{-5mm}
\vspace{-2mm}\\
\caption{Gradients of DPO and FocalPO with a focus on different sample groups where the reward estimate is correct ($p^\gamma$, $\gamma=0.05$) or incorrect ($(1-p)^\gamma$, $\gamma=1$). }
\label{fig:gradient_focus_correct_incorrect}
\vspace{-2mm}
\end{figure}

\paragraph{Effectiveness of FocalPO} 
Table~\ref{tab:main_result} demonstrates that FocalPO surpasses DPO and its variants on Alpaca Eval 2.0 and Arena-Hard. 
Notably, FocalPO improves upon SimPO, a highly performant alignment algorithm, without removing the reference model or additionally adding a sequence length control factor.
On Arena-Hard, FocalPO also shows competitive performance compared to other variants, achieving a the best score with Llama-3-Instruct and Mistral-Base-7B. 
We also present a qualitative analysis comparing outcomes between DPO and FocalPO, detailed in Appendix~\ref{appx:examples}. 
Besides, the analysis of downstream task evaluation is provided in Appendix~\ref{appx:openllm}. 


\paragraph{Focus learning correct/incorrect samples} 
FocalPO down-weighs incorrectly ranked instances that the model struggles to learn, and placing greater emphasis on correct ones compared to DPO.
To examine the impact of this choice, we empirically compare two FocalPO settings against DPO: emphasize learning \emph{incorrect} samples using $\gamma=1$ in the factor $(1-p)^{\gamma}$; and emphasize learning correct samples using $\gamma=0.05$ in $p^{\gamma}$.  
Corresponding gradient curves are illustrated in Fig.~\ref{fig:gradient_focus_correct_incorrect}. 
For a fair comparison, DPO is evaluated using its official hyperparameter setting while the learning rate is kept consistent across models focusing on incorrect and correct samples.
We present the results in Fig.~\ref{fig:focus_correct_incorrect}. We find that focusing on incorrect samples yields inferior performance compared to DPO, while focusing on correct samples outperforms DPO.
This result underscores the effectiveness our method, which prioritizes on learning preference pairs with accurate (i.e., correct) implicit reward estimates.




\section{Conclusion} 
In this paper, we propose FocalPO, a variant of DPO that dynamically scales the loss function to de-emphasize learning response pairs the implicit reward model struggles to rank, and to enhance model's understanding of pairs it can already rank correctly.
With the introduced hyperparameter fixed, we evaluate FocalPO on the most widely used chat benchmarks (Alpaca Eval 2.0 and Arena-Hard) using two models (Llama-3 and Mistral) and show that FocalPO-trained models achieve the best performance in nearly all cases. 

\section*{Limitations} 
\paragraph{More model family and hyperparameter settings}
As research line on preference optimization advances, a growing number of model-by-preference-dataset combination training settings and evaluation benchmarks have emerged.
We did not conduct experiments on all of these settings.
Besides, we only conduct experiments with~\emph{fixed} introduced hyperparameter $\gamma$ in FocalPO, limited by compute resources. 
We believe a thorough hyperparameter search could further improve the result. 
\paragraph{Future Work} the integration of our proposed FocalPO into other types of offline preference optimization loss functions remains unexplored.
For instance, SimPO~\citep{simpo} demonstrates that removing reference policy and adding sequence length as a normalization factor significantly enhance the optimization performance. 
Our experimental results show that FocalPO outperforms SimPO even without these adjustments, leaving space for the investigation of the potential benefits of combining FocalPO with these factors.

\section*{Ethics statement}
In this work, we focus on improving the effectiveness of preference learning algorithms to generate more helpful, coherent, and informative responses.
Our experiments were based on publicly available models and alignment datasets.
Despite our efforts to carefully choose our training data/model backbone, some malicious or harmful content may persist, leading to non-ideal model responses. To address this, we recognize the need to incorporate additional safety-focused datasets and commit to conducting more comprehensive evaluations on safety, harmfulness, and bias to improve the robustness and ethical alignment of future models.
We do not condone the usage of our methods/models for any unlawful or morally unjust purposes.
\bibliography{main}
\appendix

\newpage
\onecolumn

\section{Relation between FocalPO, DPO and entropy regularization}
\label{appx:relation}



\begin{lemma}
\label{lemma:1}
\emph{(Diminished entropy regularization in FocalPO)}
With $\gamma \in [0,1]$, FocalPO loss is bounded by a $\gamma$-weighted combination of the entropy of preference probability $\mathbb{H}[p(y_w \succ y_l \mid x)]$ and DPO loss $\mathcal{L}_{DPO}$: 
\begin{align}
\mathcal{L}_{\mathrm{FocalPO}} 
\leq \gamma \mathbb{H}[p(y_w \succ y_l \mid x)] + (1-\gamma) \mathcal{L}_{DPO}
\label{eq:11}
\end{align}
\end{lemma}

\textit{Proof.} 
Let $p(y_{w} \succ y_{l}|x)$ denote the optimized preference probability and $q(y_{w} \succ y_{l}|x)$ denote the target preference probability, which is not necessarily a Delta distribution, the FocalPO loss is defined as $\mathcal{L}_{\mathrm{FocalPO}} = -\mathbb{E}_{\left(x, y_w, y_l\right) \sim \mathcal{D}}
\left[p\left(y_{w} \succ y_{l} \mid x \right)^{\gamma} 
q(y_{w} \succ y_{l}|x) \log p\left(y_{w} \succ y_{l} \mid x \right)\right]$, where $\gamma, p, q \in [0, 1]$. 

Since $f(x)=x^{\gamma}$ is concave for $x\in [0,1]$, we have: 
\begin{align} 
f(x) \leq f(1) + f^{'}(1)(x-1) = \gamma x+(1-\gamma). 
\label{eq:12}
\end{align} 

Substituting this into $\mathcal{L}_{\mathrm{FocalPO}}$, we obtain:  
\begin{subequations}
\begin{align} 
\mathcal{L}_{\mathrm{FocalPO}}
\leq  &-\mathbb{E}_{\left(x, y_w, y_l\right) \sim \mathcal{D}} 
\left[[\gamma p\left(y_{w} \succ y_{l} \mid x \right)+(1-\gamma)] 
q(y_{w} \succ y_{l}|x) \log p\left(y_{w} \succ y_{l} \mid x \right)\right] \tag{13a}\\ 
= & 
\begin{aligned}[t]
&-\gamma \mathbb{E}_{\left(x, y_w, y_l\right) \sim \mathcal{D}} 
[p\left(y_{w} \succ y_{l} \mid x \right) q(y_{w} \succ y_{l}|x) \log p\left(y_{w} \succ y_{l} \mid x \right)] \\
&-(1-\gamma)\mathbb{E}_{\left(x, y_w, y_l\right) \sim \mathcal{D}} [q(y_{w} \succ y_{l}|x) \log p\left(y_{w} \succ y_{l} \mid x \right)] 
\end{aligned} \tag{13b} \label{eq_13b}
\\ 
\leq & 
\begin{aligned}[t] 
& -\gamma \max_{\left(x, y_w, y_l\right) \sim \mathcal{D}}  q(y_{w} \succ y_{l}|x) \mathbb{E}_{\left(x, y_w, y_l\right) \sim \mathcal{D}} 
[p\left(y_{w} \succ y_{l} \mid x \right) \log p\left(y_{w} \succ y_{l} \mid x \right)] \\
&+(1-\gamma) \mathcal{L}_{\mathrm{DPO}} 
\end{aligned} \tag{13c} \label{eq_13c}
\\
\leq &\gamma \mathbb{H}[p(y_w \succ y_l \mid x)] + (1-\gamma) \mathcal{L}_{DPO} \tag{13d}, 
\label{eq:11}
\end{align}
\end{subequations}
where for the step from Eq. (\ref{eq_13b}) to Eq. (\ref{eq_13c}) we use Hölder's inequality and note: 
\begin{equation}
\begin{aligned}
-&\mathbb{E}_{\left(x, y_w, y_l\right) \sim \mathcal{D}} 
[p\left(y_{w} \succ y_{l} \mid x \right) q(y_{w} \succ y_{l}|x) \log p\left(y_{w} \succ y_{l} \mid x \right)] \\
= &\left|\mathbb{E}_{\left(x, y_w, y_l\right) \sim \mathcal{D}} 
[p\left(y_{w} \succ y_{l} \mid x \right) q(y_{w} \succ y_{l}|x) \log p\left(y_{w} \succ y_{l} \mid x \right)]\right|.  
\end{aligned}    
\end{equation} 

This completes the proof.

Lemma~\ref{lemma:1} shows that the FocalPO loss is bounded by a linear combination of the preference probability entropy $\mathbb{H}[p(y_w \succ y_l \mid x)]$ and DPO loss $\mathcal{L}_{DPO}$. 
At first glance, this might resemble the entropy regularization effect in the classic focal loss~\citep{LinGGHD20}. 
In contrast, however, our motivation and experiments reveal that FocalPO actually \textbf{diminishes} the implicit \textbf{maximum-entropy regulariser} in DPO by our introduced adaptations that selectively focus on the correct cases. 
In fact, FocalPO leads to a \textbf{smoother} preference probability distribution, and curbs the model’s tendency to become overconfident in incorrect rankings in DPO. 
That said, by slightly increasing entropy, it preserves a level of uncertainty to handle data outside the preference optimization training distribution effectively.


\section{Derivation of FocalPO gradient}
\label{appx:gradient}
\begin{equation}
\begin{aligned}
\nabla_{\theta} \mathcal{L}_{\mathrm{FocalPO}} 
= \mathbb{E}_{\left(x, y_w, y_l\right) \sim \mathcal{D}}
[\nabla_{\theta} p^{\gamma}(y_{w} \succ y_{l}|x) \cdot \log & p\left(y_{w} \succ y_{l} \mid x\right) 
+ p^{\gamma}(y_{w} \succ y_{l}|x) \\
&\cdot \nabla_{\theta} \log p\left(y_{w} \succ y_{l} \mid x\right)] \\ 
\end{aligned}
\label{eq_7}
\end{equation}

Given that: 
\begin{equation}
\begin{aligned}
&\nabla_{\theta} p(y_{w} \succ y_{l}|x) 
= \nabla_{\theta} \sigma(\hat{r}_{\theta}(y_w)-\hat{r}_{\theta}(y_l)) \\
= &\beta \nabla_{\theta} \sigma(\log \frac{\pi_{\theta}(y_{w}|x)}{\pi_{ref}(y_{w}|x)} - \log \frac{\pi_{\theta}(y_{l}|x)}{\pi_{ref}(y_{l}|x)}) \\
= & \beta \nabla_{\theta}\frac{\log \pi_{\theta}(y_{w}|x)}{\log \pi_{\theta}(y_{l}|x)} \cdot \sigma(\hat{r}_{\theta}(y_w)-\hat{r}_{\theta}(y_l)) \cdot \sigma(\hat{r}_{\theta}(y_l)-\hat{r}_{\theta}(y_w)), 
\end{aligned}
\end{equation} 
where $\hat{r}_{\theta}(y)=\beta \log \frac{\pi_{\theta}(y|x)}{\pi_{ref}(y|x)}$, 
and 
\begin{equation}
\begin{aligned}
&\nabla_{\theta} \log \sigma(y) = \sigma(-y)\nabla_{\theta}y,  
\end{aligned}
\end{equation} 

Eq.~\ref{eq_7} becomes: 
\begin{equation}
\begin{aligned}
\nabla_{\theta} \mathcal{L}_{\mathrm{FocalPO}} 
&= \mathbb{E}_{\left(x, y_w, y_l\right) \sim \mathcal{D}} 
[\gamma p^{\gamma-1}(y_{w} \succ y_{l}|x) \log p\left(y_{w} \succ y_{l} \mid x\right) \cdot \nabla_{\theta}p(y_{w} \succ y_{l}|x) \\ 
&+ p^{\gamma}(y_{w} \succ y_{l}|x) \sigma(\hat{r}_{\theta}(y_l)-\hat{r}_{\theta}(y_w)) \cdot \nabla_{\theta}(\hat{r}_{\theta}(y_l)-\hat{r}_{\theta}(y_w))] \\
&=\mathbb{E}_{\left(x, y_w, y_l\right) \sim \mathcal{D}} 
[\gamma p^{\gamma-1}(y_{w} \succ y_{l}|x) \log p\left(y_{w} \succ y_{l} \mid x\right) \cdot  \beta \nabla_{\theta}\frac{\log \pi_{\theta}(y_{w}|x)}{\log \pi_{\theta}(y_{l}|x)} \\ 
&\cdot \sigma(\hat{r}_{\theta}(y_w)-\hat{r}_{\theta}(y_l)) \cdot \sigma(\hat{r}_{\theta}(y_l)-\hat{r}_{\theta}(y_w))\\ 
&+ p^{\gamma}(y_{w} \succ y_{l}|x) \sigma(\hat{r}_{\theta}(y_l)-\hat{r}_{\theta}(y_w)) \cdot \beta \nabla_{\theta}\frac{\log \pi_{\theta}(y_{w}|x)}{\log \pi_{\theta}(y_{l}|x)} \\
&= -\beta \mathbb{E}_{\left(x, y_w, y_l\right) \sim \mathcal{D}} 
\left[ \left[ \textcolor{blue}{\sigma(\hat{r}_{\theta}(y_l)-\hat{r}_{\theta}(y_w))} \right. \right. \cdot \textcolor{red}{\sigma^{\gamma}(\hat{r}_{\theta}(y_w)-\hat{r}_{\theta}(y_l))} + \textcolor{ForestGreen}{\log \sigma(\hat{r}_{\theta}(y_w)-\hat{r}_{\theta}(y_l))} \\ 
& \left. \cdot \textcolor{ForestGreen}{\gamma \sigma^{\gamma}(\hat{r}_{\theta}(y_w)-\hat{r}_{\theta}(y_l))} \cdot \textcolor{ForestGreen}{\sigma(\hat{r}_{\theta}(y_l)-\hat{r}_{\theta}(y_w))} \right] \left. \cdot \nabla_{\theta} \frac{\log \pi(y_{w}|x)}{\log \pi(y_{l}|x)} \right].
\end{aligned}
\end{equation}

\section{Training details}
We perform preference optimization on two representative models, Mistral-Base (7B)~\footnote{\url{https://huggingface.co/HuggingFaceH4/mistral-7b-sft-beta}} and instruction-tuned Llama-3 (8B)~\footnote{\url{https://huggingface.co/meta-llama/Meta-Llama-3-8B-Instruct}}. 
For Mistral-Base (7B), we follow the official hyperparameters of zephyr for the implmentation of DPO and FocalPO: $\beta=0.01$, epoch as 1, batch size as 128 and learning rate as 5e-7. 
We perform the preference optimization on UltraFeedback dataset~\citep{UltraFeedback} \footnote{\url{https://huggingface.co/datasets/trl-lib/ultrafeedback_binarized}}.  
For instruction-tuned Llama-3 (8B), we follow the setting by SimPO. 
We perform the preference optimization on llama3-ultrafeedback-armorm dataset~\footnote{\url{https://huggingface.co/datasets/princeton-nlp/llama3-ultrafeedback-armorm}} using ArmoRM~\citep{wang2024interpretable}  as the reward model. 
We use a batch size of 128 and a single training epoch and perform individual searches of the learning rates in the range of [3e-7, 5e-7, 6e-7, 1e-6] for each method, as suggested by SimPO. 
We fix the hyperparameter of $\gamma$ in FocalPO to 0.05 for both models to prevent possible overfitting of hyperparameters. 
Our code is implemented based on Transformer Reinforcement Learning (TRL) library~\citep{vonwerra2022trl} and \citet{yu2024lionsempiricallyoptimizedapproach}.
The learning rate ang $gamma$ that we adopt for training FocalPO in this paper is (5e-7, 0.05) and (6e-7, 0.05) for Mistral-Base and Llama-3-Instruct models, respectively.  

All models were trained with Flash-Attention 2~\citep{Dao24_flatt} enabled, and DeepSpeed ZeRO 3~\citep{Rasley_deepspeed}. 
We used 8 NVIDIA A100/40 GPUs for all model training. 
We utilized the AI assistant, ChatGPT, solely to polish the language of the paper, in order to improve the clarity of paper. 
We also provide the Huggingface links of baseline models that we compared with in Table~\ref{tab:main_result}.  

Mistral-Base models: 
ORPO~\footnote{\url{https://huggingface.co/kaist-ai/mistral-orpo-beta}}, 
KTO~\footnote{\url{https://huggingface.co/ContextualAI/zephyr_sft_kto}}, 
and SimPO~\footnote{\url{https://huggingface.co/princeton-nlp/Mistral-7B-Base-SFT-SimPO}}.   

Llama-3-Instruct models: 
ORPO~\footnote{\url{https://huggingface.co/princeton-nlp/princeton-nlp/Llama-3-Instruct-8B-ORPO-v0.2}}, 
KTO~\footnote{\url{https://huggingface.co/princeton-nlp/Llama-3-Instruct-8B-KTO-v0.2}}, 
and SimPO~\footnote{\url{https://huggingface.co/princeton-nlp/Llama-3-Instruct-8B-SimPO-v0.2}}. 

\section{Downstream task evaluation} 
\label{appx:openllm} 
To further investigate the impact of preference optimization on downstream tasks, we evaluate SFT, DPO and FocalPO using OpenLLM leaderboard~\citep{openllm}. 
Specifically, we assess model performance on tasks including HellaSwag~\citep{zellers-etal-2019-hellaswag}, ARC~\citep{clark2018think}, TruthfulQA~\citep{truthfulqa}, WinoGrande~\citep{sakaguchi2021winogrande}, and GSM8K~\citep{cobbe2021training} and MMLU~\citep{mmlu}. 
We use the official evaluation code of OpenLLM leaderboard~\citep{eval-harness}.

Table~\ref{tab:openllm} shows that DPO improves the leaderboard average score compared to SFT, while FocalPO \textbf{further enhances} performance beyond DPO for both Llama and Mistral models. 
Notably, FocalPO on Mistral-base achieves the highest performance on tasks of truthfulness (TruthfulQA), reading comprehension and commonsense reasoning (ARC and HellaSwag). 
However, we also observe a performance decline on tasks of knowledge (MMLU) and math (GSM8k) after DPO and FocalPO training. 
This trend aligns with prior preference optimization works (e.g.,~\citet{simpo,zhou-etal-2024-wpo}), which report similar trade-offs.  
It’s important to emphasize that the goal of alignment is to make outputs more aligned with human preferences—such as \textbf{helpfulness, truthfulness, and consistency with user intent}, etc. 
In fact, previous research~\citep{Ouyang_RLHF, askell2021general} has shown that alignment can sometimes reduce general model performance, a phenomenon referred as the "alignment tax". 
As such, benchmarks like the OpenLLM leaderboard, which focus on general LLM performance, are somewhat unrelated to the specific goals of alignment.


\begin{table*}[h]
\centering
\begin{tabular}{llllllll}
\hline
& HellaSwag & ARC  & TruthfulQA & WinoGrande & GSM8k & MMLU & Average \\ \hline
\multicolumn{8}{c}{Llama-3-Instruct   (8B)} 
\\ \hline
SFT     & 58.7      & 58.3 & 51.7       & 75.9       & 75.7  & 65.7 & 68.3    
\\
ORPO    & 58.8      & 60.2 & 57.6       & 76.7       & 75.3  & 65.6 & 69.8    
\\
KTO     & 59.3      & 60.5 & 55.6       & 75.0       & 77.5  & 66.0 & 69.6    
\\
DPO     & 59.1      & 61.3 & 56.3       & 74.7       & 75.4  & 66.1 & 69.6    
\\
SimPO   & 57.6      & 64.2 & 63.9       & 74.5       & 70.7  & 65.3 & 70.1    
\\
FocalPO & 58.7      & 63.0 & 59.7       & 74.7       & 72.6  & 66.1 & 69.9    
\\ \hline
\multicolumn{8}{c}{Mistral-base   (7B)} 
\\ \hline
SFT     & 62.1      & 54.9 & 43.0       & 77.9       & 38.8  & 59.7 & 59.9    
\\
ORPO    & 63.8      & 56.5 & 47.9       & 79.3       & 43.1  & 61.5 & 62.9    
\\
KTO     & 64.3      & 59.2 & 52.5       & 79.5       & 46.6  & 59.4 & 64.0    
\\
DPO     & 64.4      & 58.3 & 53.1       & 76.7       & 31.8  & 57.4 & 60.9    
\\
SimPO   & 64.1      & 59.8 & 50.8       & 76.9       & 31.5  & 58.4 & 60.6    
\\
FocalPO & 66.7      & 61.4 & 55.0       & 78.6       & 33.0  & 59.6 & 62.7    
\\ \hline
\end{tabular}
\caption{Downstream task evaluation results on OpenLLM leaderboard for SFT, DPO and FocalPO for Mistral-Base (7B) and Llama-3-Instruct (8B). }
\label{tab:openllm}
\end{table*}

\section{Related works}
Several preference optimization loss functions have been proposed as alternatives to DPO.
For instance, KTO ~\citep{KTO} proposes to directly maximize the utility of generations instead of maximize the likelihood of preferences. 
ORPO~\citep{hong-etal-2024-orpo} focuses on jointly optimizing SFT and preference optimization loss by adding the odds ratio term. 
SimPO~\citep{simpo} removes the requirement of reference model, and introduces a target reward margin and a length-normalized factor. 
ODPO~\cite{AminiVC24ODPO} further modifies DPO to incorporates additional labeled reward values and also employs an instance-dependent margin term.
By adding an additional margin term, we find many approaches \emph{implicitly} reduces the high weights assigned to learning incorrect response pair used in the original DPO.
In this work, we propose FocalPO inspired by Focal Loss from vision tasks, to explicitly reduces emphasis on learning response pairs that the model struggles to rank while prioritizing preference pairs with accurate implicit reward estimates.

\section{Qualitative analysis}
\label{appx:examples}
We present a pair of answers generated by Llama-3-Instruct after training with DPO and FocalPO in Table~\ref{tab:example1} and \ref{tab:example2}, respectively. 
Given the question of \textit{How long will it take to walk around the world, assuming that would be possible}, Llama-3-Instruct (DPO) gives an answer of around 335 days, without considering factors like human rest time. 
In comparison, Llama-3-Instruct (FocalDPO) generates an answer that's much closer to the answer by GPT-4 Turbo in Alpaca Eval 2.0, 1,169 days, and exactly aligned with the answer by O1, 1,002 days.

\begin{table*}[h]
\centering
\begin{tabular}{ll}
\hline
Instruction & \begin{tabular}[c]{@{}l@{}}How long will it take to walk around the world, assuming that would be \\ possible?\end{tabular} 
\\ \hline
\begin{tabular}[c]{@{}l@{}}Llama-3-Instruct \\ (DPO) \end{tabular}       & \begin{tabular}[c]{@{}l@{}}What an exciting question! Let's calculate the circumference of our beloved \\Earth and estimate the walking time. Here's the breakdown: \\ \\ \textbf{Earth's Circumference:}\\ The Earth's equatorial circumference is approximately \textbf{40,075 kilometers} \\\textbf{(24,901 miles}. Since we'll assume a continuous, unobstructed path, we'll use \\this value.\\ \\ \textbf{Walking Speed:}\\ A moderate, comfortable walking pace is about \textbf{5 kilometers per hour} \textbf{(3.1} \\\textbf{miles per hour)}. Let's use this speed for our calculation.\\ \\ \textbf{Calculations:}\\ \\ 1. Convert circumference from kilometers to hours: 40,075 km ÷ 5 km/h = \\\textbf{8,015 hours}\\ 2. Convert hours to days: 8,015 hours $\approx$ \textbf{334.8 days} (or approximately 11 \\months)\\ 3. Convert days to years: \textbf{0.92 years} (assuming a non-stop, 365-day year)\\ \\ So, if you could walk around the world without stopping, it would take \\ 
approximately \textbf{0.92 years} or \textbf{11 months} to complete the journey, assuming a \\
consistent moderate pace of 5 km/h.\\ \\ Keep in mind that this calculation ignores factors like:\\ \\ * Rest breaks, meals, and sleep\\ * Terrain difficulties, like mountains, rivers, and urban areas\\ * Weather conditions, like extreme temperatures, storms, or darkness\\ * Border crossings, permits, and logistical challenges\\ \\ While walking around the world might not be feasible in reality, this \\calculation provides a fascinating estimate of the incredible distance involved. \\Happy walking (in your imagination)!\end{tabular} 
\\ \hline
\end{tabular}
\caption{An Alpaca Eval 2.0 generation from the Llama-3-Instruct model after training with DPO. }
\label{tab:example1}
\end{table*}

\begin{table*}[h]
\centering
\begin{tabular}{ll}
\hline
Instruction & \begin{tabular}[c]{@{}l@{}}How long will it take to walk around the world, assuming that would be \\ possible?\end{tabular}        
\\ \hline 
\begin{tabular}[c]{@{}l@{}}Llama-3-Instruct \\ (FocalPO) \end{tabular}  & \begin{tabular}[c]{@{}l@{}}What a fascinating question! Let's break it down to estimate the time it'd \\take to walk around the world. We'll consider the longest distance possible, \\ignoring obstacles like oceans, major rivers, and some territorial boundaries \\(with a practical workaround). \\ \\ \textbf{The approximate distance:}\\ The Earth's circumference at the equator is approximately \textbf{40,075} \\\textbf{kilometers (24,901 miles)}.\\ \\ \textbf{Assumptions:}\\ \\ 1. \textbf{Walking speed:} A moderate, consistent pace of 5 kilometers per hour (3.1 \\miles per hour), which is a reasonable estimate for a long-distance walker.\\ 2. \textbf{No rest days} (except for occasional breaks for necessities like food, water, \\and sleep). We'll assume walking continuously, 8 hours a day.\\ 3. \textbf{No altitude adjustments} (ignoring high-altitude trekking challenges).\\ \\ \textbf{Calculations:}\\ \\ 1. Daily walking distance: 5 km/h × 8 hours = 40 kilometers (24.9 miles) per \\day\\ 2. Total days required: 40,075 km (circumference) ÷ 40 km/day = \textbf{1,001.88} \\\textbf{days} or approximately \textbf{2.74 years}\\ \\ So, assuming a non-stop, moderate-paced walk around the world, it would take \\approximately \textbf{2 years and 9 months} to complete the journey. Keep in mind \\that this is a theoretical estimate, neglecting factors like:\\ \\ * Border crossings and route adjustments\\ * Terrain difficulties (mountains, deserts, urban areas)\\ * Health considerations (injuries, fatigue, climate extremes)\\ * Logistical challenges (food, accommodation, permits)\\ \\ While it's not feasible to walk around the world without significant planning \\and preparation, this calculation gives you an idea of the incredible duration \\involved. Happy walking (in theory)!\end{tabular} 
\\ \hline
\end{tabular}
\caption{
An Alpaca Eval 2.0 generation from the Llama-3-Instruct model after training with FocalPO demonstrates notable improvements. Compared to the output generated by the DPO model (Table~\ref{tab:example1}), the output generated by FocalPO is more realistic, as it considers factors such as the time humans need to rest. 
This answer is much closer to the answer by GPT-4 Turbo in Alpaca Eval 2.0, 1,169 days, and exactly aligned with the answer by O1, 1,002 days.}
\label{tab:example2}
\end{table*}

\end{document}